\documentclass[10pt,twocolumn,letterpaper]{article}
\pdfoutput=1
\usepackage{cvpr}
\usepackage{times}
\usepackage{epsfig}
\usepackage{graphicx}
\usepackage{amsmath}
\usepackage{amssymb}
\usepackage{algorithm}
\usepackage{algorithmic}
\DeclareMathOperator*{\argmax}{arg\,max}
\usepackage{hyperref}
\usepackage{subcaption}

\cvprfinalcopy 

\def\cvprPaperID{3001} 

\ifcvprfinal\fi

\begin{document}

\title{SOS: Selective Objective Switch for Rapid Immunofluorescence Whole Slide Image Classification}

\author{Sam Maksoud$^{1,2}$ \qquad Kun Zhao$^{1}$ \qquad Peter Hobson$^{2}$ \qquad Anthony Jennings$^{2}$ \qquad Brian C. Lovell$^{1}$ \\
$^{1}$The University of Queensland, St Lucia QLD 4072, Australia\\
$^{2}$Sullivan Nicolaides Pathology, Bowen Hills, QLD 4006, Australia\\
}

\maketitle

\begin{abstract}
	The difficulty of processing gigapixel whole slide images (WSIs) in clinical microscopy has been a long-standing barrier to implementing computer aided diagnostic systems.  Since modern computing resources are unable to perform computations at this extremely large scale, current state of the art methods utilize patch-based processing to preserve the resolution of WSIs. However, these methods are often resource intensive and make significant compromises on processing time. In this paper, we demonstrate that conventional patch-based processing is redundant for certain WSI classification tasks where high resolution is only required in a minority of cases. This reflects what is observed in clinical practice; where a pathologist may screen slides using a low power objective and only switch to a high power in cases where they are uncertain about their findings. To eliminate these redundancies, we propose a method for the selective use of high resolution processing based on the confidence of predictions on downscaled WSIs --- we call this the Selective Objective Switch (SOS). Our method is validated on a novel dataset of 684 Liver-Kidney-Stomach immunofluorescence WSIs routinely used in the investigation of autoimmune liver disease. By limiting high resolution processing to cases which cannot be classified confidently at low resolution, we maintain the accuracy of patch-level analysis whilst reducing the inference time by a factor of 7.74.
	
\end{abstract}

\section{Introduction}

\begin{figure}[t]
\begin{center}
   \includegraphics[width=0.7\linewidth]{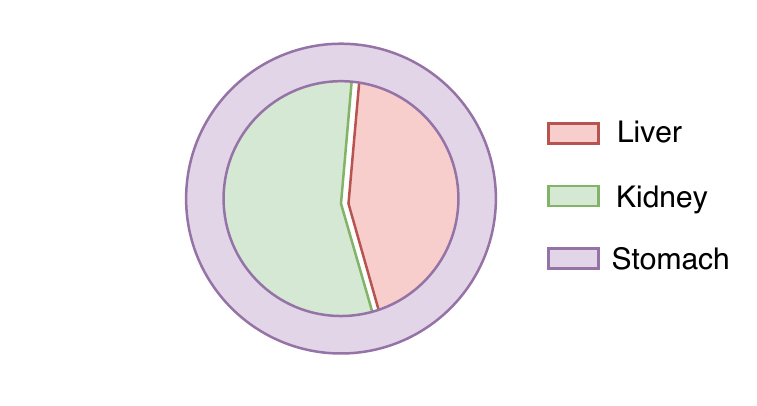}
\end{center}
   \caption{Schematic illustration of a WSI that includes multi-tissue types: liver, kidney and stomach.}
\label{fig:lks}
\end{figure}

Patch-level image processing with convolutional neural networks (CNNs) is arguably the most widely used method for gigapixel whole slide image (WSI) analysis \cite{popular}. Ordinarily, processing such a large image in its totality with CNNs is computationally infeasible without significant downscaling --- resulting in the loss of detailed information required for fine-grained analytical tasks. However, by processing WSIs in smaller patches, it is possible to extract the detailed information by preserving the resolution of the original gigapixel image. Thus, applications that require fine-grained analysis of gigapixel WSIs are able to incorporate powerful CNNs in their design.

Although there are clear advantages to high resolution patch-level analysis \cite{patch_em}, it is resource intensive and can substantially increase processing time \cite{slow}. In a high throughput laboratory, any additional per sample processing time will compound, which can make it difficult to justify the use of deep learning algorithms for WSI analysis. Hence, there is a strong motivation to identify situations where it is unnecessary to process WSIs at their maximum resolution. Unlike bright field microscopy, the increased sensitivity of the immunofluorescence assay allows for analysis at lower resolutions. Indirect immunofluorescence (IIF) microscopy on multi-tissue sections is one such example where low resolution often provides sufficient discriminatory information.

Multi-tissue IIF slides, such as the Liver-Kidney-Stomach (LKS) slide shown in Figure \ref{fig:lks}, allow for the simultaneous observation of immunoreactivity across different tissue types. Comparing observations across tissue types is crucial to interpreting these WSIs \cite{ald}; so it is advantageous to screen them at a lower magnification which allows multiple tissues to be viewed at once. Most patterns are easily
identified at these lower magnifications. Accordingly, a pathologist will often only switch to a higher magnification for complex or ambiguous cases that require a greater resolving power. Conventional patch-based processing methods do not reflect this highly efficient manner in which humans navigate slides in clinical microscopy.

In this paper, we describe an approach to WSI classification using a mechanism which restricts the use of high resolution processing to the complex or ambiguous cases. To this end, we construct a dynamic multi-scale WSI classification system comprising three key components: a Low Resolution Network (LRN); an Executive Processing Unit (EPU); and a High Resolution Network (HRN). Inspired by the efficient screening techniques used in manual IIF microscopy, we first attempt to classify WSIs with low resolution features extracted from the LRN. The EPU triggers high resolution patch-based processing iff the probability of the class predicted at low resolution is below a prescribed confidence threshold. We refer to this protocol as the Selective Objective Switch (SOS). The contributions of this paper are as follows:

\begin{itemize}
  \item To our knowledge, we are the first to propose a Dynamic Multi-Scale WSI classification network which regulates the use of high resolution image streams via the uncertainty of predictions at low resolution;
  \item We introduce a novel learning constraint, the paradoxical loss, to discourage asynchronous optimization of the LRN and HRN during training; 

  \item Finally, we will release our novel dataset\footnote{\href{https://github.com/cradleai/LKS-Dataset}{https://github.com/cradleai/LKS-Dataset}} of $684$ LKS WSIs to the community. This will be the first publicly available dataset for multi-tissue IIF WSI analysis.  
\end{itemize}

\section{Related Works}
\label{sect:related}

The current methods used for WSI analysis can be broadly classed into patch-level, conventional multi-scale, and dynamic multi-scale image processing.

\textbf{Patch-Level Methods}. To classify a WSI from patches, patch-level CNNs must incorporate a decision or feature fusion method to aggregate the information from multiple image patch sources. In \cite{patch_pretrain}, Xu \textit{et al}. use a 3-norm pooling method to aggregate patch-level features, extracted from a CNN pre-trained on ImageNet \cite{imagenet}, prior to classification. While this method was able to outperform image-level classification of Low-Grade Glioma (LGG) by a significant margin, Hou \textit{et al}. \cite{patch_em} later discovered that pooling general patch features does not capture the heterogeneity that differentiates subtypes of LGG. This suggests that the method described in \cite{patch_pretrain} is not suitable for fine-grained WSI classification tasks. Hou \textit{et al}.  were able to achieve fine-grained classification of LGG subtypes by applying fine-tuning to the CNNs during training and deriving WSI classifications from aggregated predictions on individual patches \cite{patch_em}. However, this assumes that a WSI can be classified based on observations made in a single patch --- making it unsuitable for classification tasks that require correlating features from multiple locations in a WSI.


\textbf{Conventional Multi-Scale Methods}. Multi-scale networks provide a means of capturing spatial context in WSIs without compromising on detail. Due to the small receptive field, a single WSI patch may contain little to no contextual information \cite{patch_context, patch_adaptive}. This is not a major hindrance to cancer classification on many popular datasets \cite{tcga, bach, camelyon17, camelyon16}; as these WSIs are classified based on cellular mutations observable at the patch-level. However, for tasks that require analysis of a broader WSI context, capturing long range spatial dependencies is of vital importance \cite{patch_context, patch_lstm,  patch_adaptive}.

An obvious way to capture long range dependencies in CNNs would be to increase the size of input patches as described by Pinheiro and Collobert \cite{sem_seg1}. However, in the case of gigapixel WSIs, complex long range dependencies may span across tens of thousands of pixels. Without downsampling, capturing them with larger input patches is computationally impossible. Multi-scale networks resolve this problem by using multiple input streams to capture different levels of detail \cite{patch_context, multiscale_pyramid, patch_lstm}.

Ghafoorian \textit{et al}. \cite{multiscale_pyramid} proposed a multi-scale late fusion pyramid architecture where low resolution image streams with a large field of view (FOV) were used to capture spatial context, while high resolution image streams with a smaller FOV captured the finer details. Although this approach is effective at capturing different levels of detail, Sirinukunwattana \textit{et al}. \cite{patch_lstm} showed that using long short-term memory (LSTM) units \cite{lstm} to integrate features from multiple scales is more robust to noise, less sensitive to the order of inputs and generally more accurate than the late fusion method used by Ghafoorian \textit{et al}. \cite{multiscale_pyramid}. While both of these multi-scale approaches perform better than traditional single-scale patch-level methods \cite{patch_lstm}, incorporating additional image data at different resolutions increases the computational cost of WSI analysis.


\begin{figure*}[t]
\begin{center}
   \includegraphics[width=0.9\linewidth]{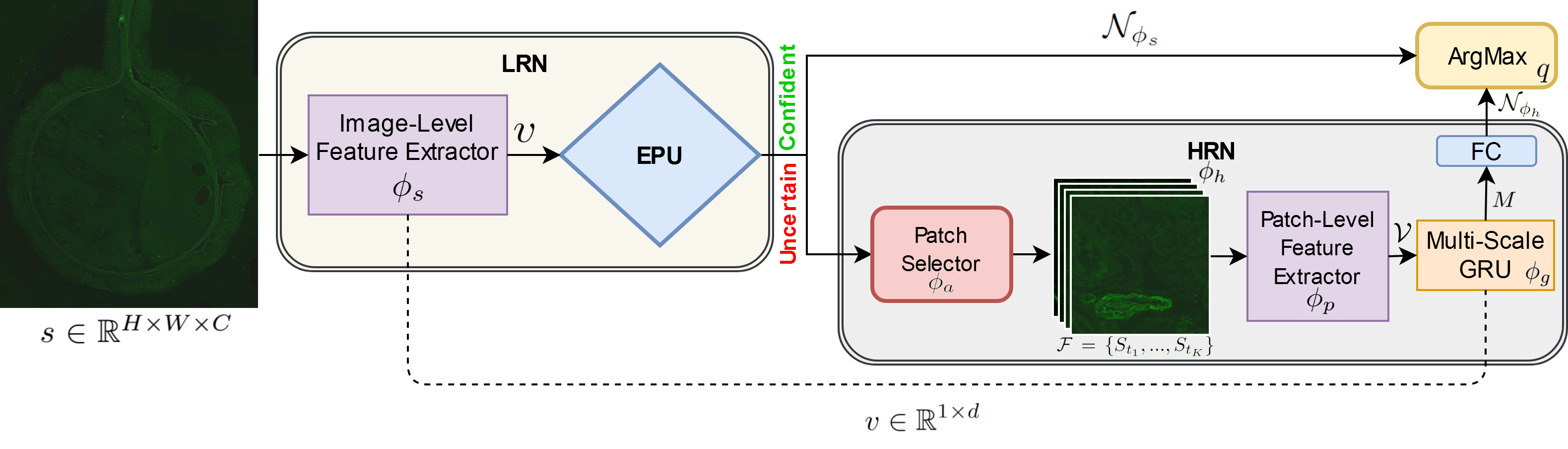}
\end{center}
   \caption{Framework of the SOS protocol. Dashed lines indicate the residual connection between the LRN and HRN.}
\label{fig:ionet}
\end{figure*}

\textbf{Dynamic Multi-Scale Methods}. The major disadvantage of using conventional multi-scale methods is the overwhelming redundancies in the visual information fed into these systems. In this paper, we refer to techniques that regulate the degree of information received from different image scales as dynamic multi-scale methods.

Excessive fixation on diagnostically irrelevant WSI features is thought to be the reason why novice pathologists are considerably slower and less accurate than experts; who direct their attention to highly discriminative regions \cite{eye_study}. BenTaib and Hamarneh \cite{patch_context} showed that multi-scale networks behave in a similar manner. Using recurrent visual attention, they outperform classification models trained on thousands of patches by selecting only 42 highly discriminate patches at various scales. Similar improvements are observed in pixel-wise WSI classification tasks. Tokunaga \textit{et al}. \cite{patch_adaptive} found that subtypes of lung adenocarcinoma have different optimal resolutions for observing discriminatory features. By dynamically adapting the weight of features from multiple image scales, they could focus on the most discriminative features for detecting the type of cancer lesion. While both of these methods are capable of adapting to the most relevant features in an image, neither adjust the number of patches used, and hence, processing time \cite{slow}, to suit the individual requirements of each WSI.

The most similar work to ours is from Dong \textit{et al}. \cite{razn}, who proposed training a policy network \cite{policy} based on the ResNet18 architecture \cite{resnet} to decide whether to use high or low resolution image scales to process a WSI. Although this resulted in faster processing time for WSI segmentation tasks \cite{razn}, there are several disadvantages of using this approach for WSI classification. Firstly, training a separate policy network to decide on which image scale to use introduces a significant number of model parameters. In contrast, our decision protocol is based on the predictive confidence at low resolution, thus avoiding the redundant first pass through a policy network. Secondly, their high resolution pathway does not incorporate features from low resolution image scales. In effect, the ability to capture long range spatial dependencies is significantly impaired \cite{multiscale_pyramid, patch_lstm}. However, we overcome this problem by recycling feature maps from the low resolution pathway to incorporate spatial context information. Finally, even though Dong \textit{et al} computed the error for both the low resolution and high resolution image scale pathways during training, they only updated the parameters of a single pathway for each instance --- which is computationally wasteful. In our method, all image scale pathways are updated for each training instance.

\section{Selective Objective Switch}
As illustrated in Figure~\ref{fig:ionet}, the aim of the SOS protocol is to avoid excessive high resolution patch-level processing for WSIs that can be classified confidently at the image-level. To this end, we train an EPU to serve as a controller that decides whether the LRN or HRN is used to classify a given WSI. We describe the details of these components and our optimization protocol below.

\subsection{Model Framework}
\textbf{Low Resolution Network.} Depending on the path chosen by the EPU, the LRN can serve as either a WSI classifier, or a feature extractor for the HRN. The LRN receives a downscaled WSI, $s \in \mathbb{R}^{H\times W\times C}$, as input to a ResNet18 based feature extractor, $\phi_{s}$, to compute a high level feature vector $v$ as:
\begin{equation}
\label{eq:lrt_feats}
\phi_{s}(s) = v \in \mathbb{R}^{1\times d},
\end{equation}
where $d=512$ is the number of output channels from the penultimate layer of ResNet18 \cite{resnet}.

\textbf{Executive Processing Unit.} The EPU is located at the terminal end of the LRN where it receives $v$ as input and estimates a set of class probabilities $\mathcal{N}_{\phi_{s}} = \{ {N_{s}}_{1},...,{N_{s}}_{n}\}$, where $n$ is number of WSI classes. To compute  $\mathcal{N}_{\phi_{s}}$, we apply a linear transformation to $v$ followed by the softmax function $\sigma$:
\begin{equation}
\label{eq:lrt_pred}
\mathcal{N}_{\phi_{s}} = \sigma\left(vA_{s}^{T} + b_{s}\right),
\end{equation}
where $A_{s} \in \mathbb{R}^{n\times d}$ and $b_{s} \in \mathbb{R}^{n}$ are parameters learned by the network. The element with the highest value in $\mathcal{N}_{\phi_{s}}$ is compared to a confidence threshold $c$ in the range $[0,1]$ to determine the flow of downstream operations.

\begin{algorithm}
\caption{EPU Switch Statement}
\label{code:switch}
\begin{algorithmic}[1]
\IF {$\max\left(\mathcal{N}_{\phi_{s}}\right) > c$}
    \STATE $q = \argmax \left(\mathcal{N}_{\phi_{s}}\right)$
\ELSE
	\STATE $q = \argmax \left(\phi_{h}(v)\right)$
\ENDIF
\end{algorithmic}
\end{algorithm}
As shown in Algorithm \ref{code:switch}, the high confidence estimations immediately compute the class label $q$ using the $\argmax$ function while the low confidence estimations trigger additional processing by the HRN $\phi_{h}$. The details of the $\phi_{h}$ function are outlined below.

\begin{figure}[t]
\begin{center}
   \includegraphics[width=0.6\linewidth]{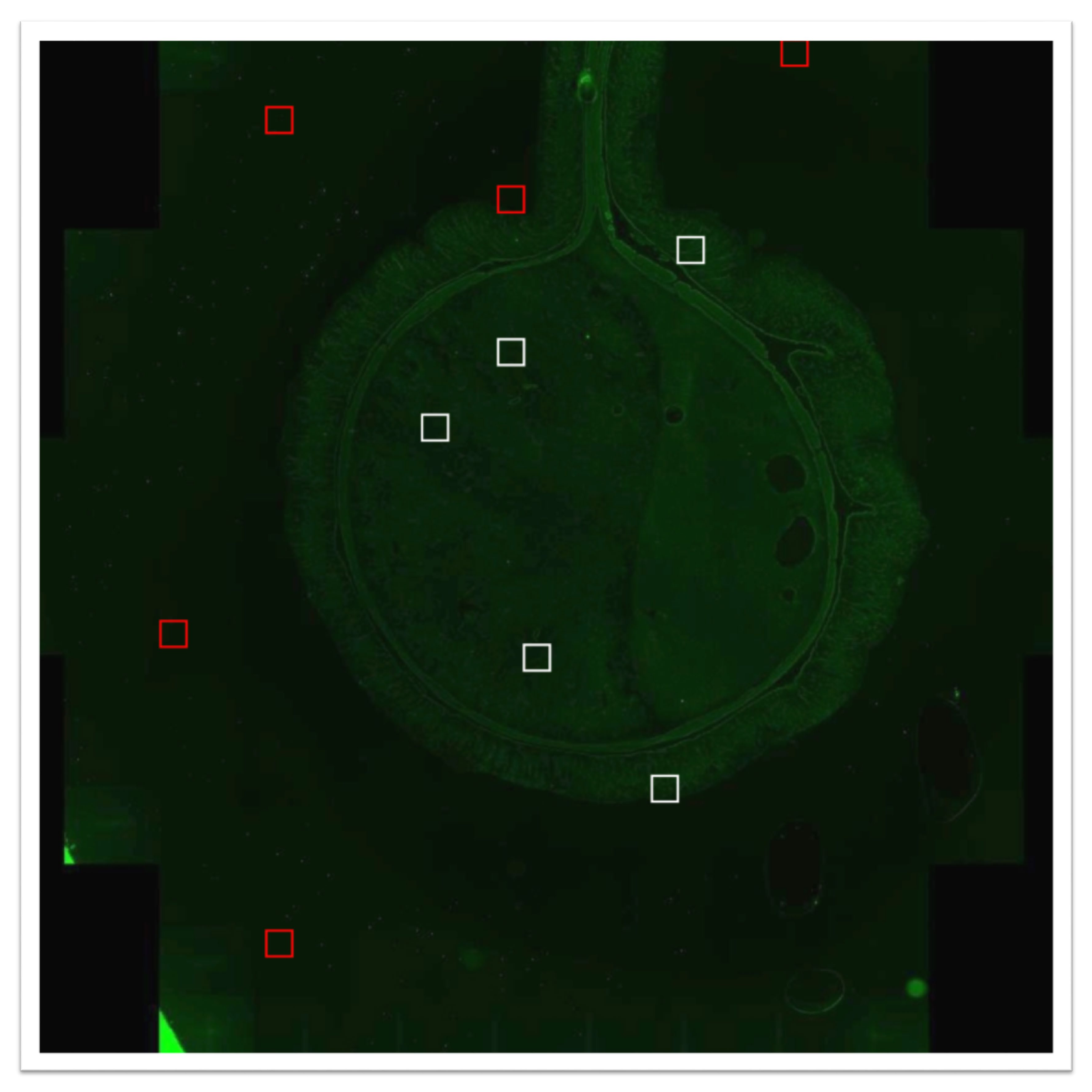}
\end{center}
   \caption{Visualization of patch selection in the HRN. Red boxes indicate undesirable selections of non-tissue regions. }
\label{fig:bad}
\end{figure}

\textbf{High Resolution Network.} The HRN, $\phi_{h}$, comprises three main subcomponents; a patch selector $\phi_{a}$, a patch-level feature extractor $\phi_{p}$, and a Gated Recurrent Unit (GRU) $\phi_{g}$  \cite{gru}. The patch selector function $\phi_{a}$ is based on the stochastic hard attention mechanism proposed by Xu \textit{et al}. \cite{attention}. Specifically, the indices of elements in $\mathcal{S} = \{ S_{1},...,S_{P}\}, S_{p} \in \mathbb{R}^{H\times W\times C}$ are treated as intermediate latent variables where $\mathcal{S}$ is the set of $P$ high resolution patches derived from the full resolution WSI. We then estimate a Multinoulli distribution $\mathcal{X}$ as a function of the image-level WSI features $v$:

\begin{equation}
\label{eq:attention}
\mathcal{X} = \sigma\left(vA_{p}^{T} + b_{p}\right),
\end{equation}
where $A_{p} \in \mathbb{R}^{P\times d}$ and $b_{p} \in \mathbb{R}^{P}$ are parameters learned by the network. The indices of the $K$ highest elements in $\mathcal{X}$ are used to sample the set of discriminate patches $\mathcal{F} = \{ S_{t_{1}},...,S_{t_{K}}\}, S_{t_{k}} \in \mathcal{S}$. 
The value of $K$ is limited by the maximum capacity of the GPU. In our experiments the upper limit of $K$ is $10$. A set of patch-level feature vectors $\mathcal{V} = \{ V_{1},...,V_{K}\}$ are then extracted by applying the $\phi_{p}$ function to each element in $\mathcal{F}$. The architecture of $\phi_{p}$ is a clone of $\phi_{s}$. The reason for using a separate network to extract features from the high resolution patches is that CNNs are not robust to changes in scales \cite{snip}. Thus, the objective of $\phi_{p}$ becomes:

\begin{equation}
\label{eq:hrt_feats}
\phi_{p}(\mathcal{F}) = \{ V_{1},...,V_{K}\}, V_{p} \in \mathbb{R}^{1\times d}.
\end{equation}

The $\phi_{g}$ function receives $\mathcal{V}$ and $v$ (via the residual connection to $\phi_{s}$ illustrated in Figure \ref{fig:ionet}) as input and computes $M \in \mathbb{R}^{2d}$, where $M$ is a multi-scale representation of the WSI. The design of $\phi_{g}$ is similar to the late fusion multi-stream LSTM architecture described in \cite{patch_lstm}, however, we substitute the LSTM for a GRU cell as they have been observed to achieve comparable performance at a lower computational cost \cite{gru_vs_lstm}. The GRU cell, with hidden state $h \in \mathbb{R}^{d}$, is first initialized with the image-level feature vector ($v$), and subsequently receives a patch image feature vector in $\mathcal{V}$ each time step for a total of $K$ time steps. The final state of the GRU cell is concatenated with $v$ to construct $M$. Finally, we compute the class label $q$ as follows:
\begin{equation}
\label{eq:hrt_preds}
\mathcal{N}_{\phi_{h}} = \sigma\left(MA_{m}^{T} + b_{m}\right),
\end{equation}
\begin{equation}
\label{eq:hrt_class}
q = \argmax\left(\mathcal{N}_{\phi_{h}}\right),
\end{equation}
where $A_{m} \in \mathbb{R}^{n\times 2d}$ and $b_{m} \in \mathbb{R}^{n}$ are parameters learned by the network and $\mathcal{N}_{\phi_{h}} = \{ {N_{h}}_{1},...,{N_{h}}_{n}\}$ is the set of estimated class probabilities.

\subsection{Optimization Protocol}
Our model is trained by optimizing three loss terms: classification loss; paradoxical loss; and executive loss. During training, the EPU always triggers HRN processing to optimize the classification accuracy of both networks. During inference, the EPU uses the switch statement in Algorithm~\ref{code:switch} to decide on a single path. We describe the details of our optimization protocol below.

\textbf{Classification Loss.} The classification loss $L_{1}$ is the summation of two cross entropy loss terms: a low resolution cross entropy loss $L_{ce_{1}}$; and a high resolution cross entropy loss $L_{ce_{2}}$. The purpose of $L_{ce_{1}}$ is to maximize the classification accuracy when inferring the class label $q$ from the LRN probability distribution $\mathcal{N}_{\phi_{s}}$:

\begin{equation}
\label{eq:LRCE}
L_{ce_{1}} = \frac{1}{B} \sum^{B}_{o=1} \left(-\sum^{n}_{i=1} y_{o,i}\log({N_{s}}_{o,i})\right),
\end{equation}

where $B=4$ is the mini-batch size, $i$ is the class label, $o$ is the observed WSI, $y$ is a binary indication that $i$ is the ground truth label for $o$, and ${N_{s}}_{o,i}$ is the probability that $o = i$ if predicted by the LRN. The purpose of $L_{ce_{2}}$ is to maximize the classification accuracy when inferring $q$ from the HRN probability distribution $\mathcal{N}_{\phi_{h}}$. The $L_{ce_{2}}$ term is the same as Equation \ref{eq:LRCE}, except we use probabilities in $\mathcal{N}_{\phi_{h}}$ instead of $\mathcal{N}_{\phi_{s}}$ to calculate the cross entropy loss. The $L_{1}$ loss is then computed as the sum of $L_{ce_{1}}$ and $L_{ce_{2}}$.

\textbf{Paradoxical Loss.} The motivation of the paradoxical loss $L_{2}$ term is the assumption that, given $M$ is a multi-scale representation of image-level and patch-level features, access to more visual detail should never decrease the performance of the HRN. Thus, instances where the LRN performs better than the HRN during training should be viewed as undesirable and paradoxical. Under this assumption, we hypothesize that if the probability of the correct class is higher in $\mathcal{N}_{\phi_{s}}$ than in $\mathcal{N}_{\phi_{h}}$, it must be due to either an overconfident LRN, or a suboptimal HRN. To deter these behaviors in our model we compute $L_{2}$ as follows:
\begin{equation}
\label{eq:L2}
L_{2} = \frac{1}{B} \sum^{B}_{o=1} \max\left({N_{s}}_{x, o} - {N_{h}}_{x, o}, 0\right),
\end{equation}
where ${N_{s}}_{x, o}$ and ${N_{h}}_{x, o}$ are the estimated probabilities of the true class label $x$ by the LRN and HRN respectively.

\textbf{Executive Loss.} The Executive Loss $L_{3}$ is a weighted sum of two novel loss terms: a hesitation loss; and a hubristic loss. Its purpose is to calibrate both the confidence threshold $c$ (Algorithm \ref{code:switch}) and the LRN confidence scores to achieve the optimal trade-off between efficiency and accuracy. This is crucial because confidence scores naturally produced by neural networks may not represent true probabilities \cite{cal}. Intuitively, the hesitation loss and hubristic loss can be understood as the difference between the predicted probability value, and the value relative to $c$ that would have resulted in a correct action by the EPU.

The hesitation loss, $L_{he}$, is the penalty incurred when there is a high degree of uncertainty in correct LRN predictions, resulting in redundant HRN processing. Specifically, this describes instances when: (a) the LRN predicts the correct class label, and (b) the predicted probability value is below the confidence threshold. To prevent our network from using the HRN excessively, we penalize correct LRN predictions when the probability value is below $c$ by computing $L_{he}$ as follows:
\begin{equation}
\label{eq:hesitation}
L_{he} = \sum^{B}_{o=1} y_{s, o} \max\left(\left((c + \epsilon) - \max\left(\mathcal{N}_{\phi_{s}}\right)\right), 0 \right),
\end{equation}
where $\epsilon = 10^{-3}$ is used to set the desired target above the confidence threshold, and $y_{s, o}$ is the binary indicator that $\argmax\left(\mathcal{N}_{\phi_{s}}\right)$ is the correct label for observation $o$.

The hubristic loss $L_{hu}$ is the penalty incurred when the EPU's decision to bypass correct HRN predictions is based on confidently incorrect predictions by the LRN. Specifically, this describes instances when: (a) the LRN predicts an incorrect class label; (b) the predicted probability value is above the confidence threshold; and (c) the HRN predicts the correct class label. To prevent this underutilization of the HRN, we penalize incorrect LRN predictions that are predicted with a probability value above $c$ by computing $L_{hu}$ as:
\begin{equation}
\label{eq:hubristic}
L_{hu} = \sum^{B}_{o=1} y_{h, o} \neg (y_{s, o}) \max\left(\left(\max\left(\mathcal{N}_{\phi_{s}}\right)\right) - c, 0 \right),
\end{equation}
where $y_{h, o}$ is binary indicator that $\argmax\left(\mathcal{N}_{\phi_{h}}\right)$ is the correct class label for $o$, and $\neg (y_{s, o})$ is the binary indicator that $\argmax\left(\mathcal{N}_{\phi_{s}}\right)$ is the incorrect class label for $o$.

Both $L_{he}$ and $L_{hu}$ are then weighted to compute $L_{3}$ as:
\begin{equation}
\label{eq:L3}
L_{3} = \frac{1}{B} \left(\lambda_{1}L_{he} + \lambda_{2}L_{hu}\right),
\end{equation}
where $\lambda_{1}=0.5$ and $\lambda_{2}=1.0$ are regularization terms that set the target speed/accuracy trade-off by controlling the influence of $L_{he}$ and $L_{hu}$ on $L_{3}$.

\textbf{Final Objective Function.} The final objective function $L_{total}$ is computed as the sum of $L_{1}$, $L_{2}$, and $L_{3}$ terms. By adding $L_{3}$ to the total network loss (rather than simply calibrating the threshold $c$), our network has the flexibility to improve EPU decision making by: (a) adjusting $c$ directly; and/or (b) modifying other network parameters to have the LRN regress confidence scores on the appropriate side of $c$.

\section{The Liver-Kidney-Stomach Dataset}

\begin{table}
\small
\begin{subtable}{\linewidth}
\begin{center}
\begin{tabular}{c| c c c c c}
Set & Neg &  AMA & SMA-V & SMA-T & Total \\ \hline
Train & $239$ & $106$ & $107$ & $27$ & $479$ \\
Test & $103$ & $45$ & $46$ & $11$ & $205$ \\
\end{tabular}
\end{center}
\caption{The distribution of classes in the train and test set.}
\label{table:class}
\end{subtable}
\newline
\vspace*{0.2 cm}
\newline
\begin{subtable}{\linewidth}
\begin{center}
\begin{tabular}{c|c|c|c}

Size & Resolution & Objective & Format\\ \hline
300GB & $40000\times40000\times1$ & $\times20$ & TIFF
\end{tabular}
\end{center}
\caption{Meta-Information pertaining to the LKS dataset.}
\label{table:meta}
\end{subtable}
\caption{Structure of the Liver-Kidney-Stomach Dataset.}
\end{table}

The liver auto-antibody LKS screen is critical to the investigation of autoimmune liver disease \cite{ald, ald2}, however, there are currently no public WSI datasets available for research. There are several reasons why the LKS classification task is ideal for evaluating dynamic multi-scale networks. Firstly, compared to public bright field microscopy WSI datasets, the increased sensitivity of the immunoflouresence assay allows for the observation of critical features at low resolution. Secondly, despite the increased sensitivity of IIF, high resolution may still be required to observe certain staining patterns, particularly when antibody concentrations are low. Finally, global structures captured at low resolution are essential for the classification of multi-tissue LKS WSIs. The fact that low resolution features are necessary, but not always sufficient, for LKS classification provides an ideal environment to validate the SOS protocol. 

In collaboration with Sullivan Nicolaides Pathology, we constructed a novel LKS dataset from routine clinical samples. To prepare the LKS slides, sections of rodent kidney, stomach and liver tissue were prepared according to the schematic in Figure \ref{fig:lks}. Patient serum was incubated on the multi-tissue section and treated with fluorescein isothiocyanate (FITC) IgG conjugate. The slides were digitized using a monocolor camera and a x$20$ objective lens with a numerical aperture of $0.8$. 
A team of trained medical scientists manually labelled the slides into one of four classes: Negative (Neg); Anti-Mitochondrial Antibodies (AMA); Vessel-Type  Anti-Smooth Muscle Antibodies (SMA-V) and Tubule-Type Anti-Smooth Muscle Antibodies (SMA-T). The distribution of classes is provided in Table \ref{table:class} and relevant meta-information is provided in Table \ref{table:meta}. 


\section{Experiments}

\begin{table}
\small
\begin{subtable}{\linewidth}\centering
{\begin{tabular}{c c c | c  c c | c c}
$L_{2}$ & $L_{3}$ & $K$ & TA\% $\uparrow$ & IT(s) $\downarrow$ & RS $\downarrow$ & LP & CT\\ \hline
\checkmark & \checkmark & $10$ & $\bf{90.73}$ & $15.78$ & $2.17$ & $0.94$ & $0.62$ \\
\checkmark & \checkmark & $5$ & $88.29$ & $9.74$& $2.17$ & $0.92$ & $0.60$ \\
\checkmark & \checkmark & $3$ & $86.83$ & $13.15$& $2.17$ & $0.89$ & $0.61$ \\ \hline
  & \checkmark & $10$ & $85.37$ & $\bf{8.92}$ & $2.17$& $0.97$ & $0.52$ \\ 
\checkmark &  & $10$ & $88.29$ & $15.50$ & $2.17$& $0.92$ & $0.62$ \\ 
\end{tabular}}
\caption{\textbf{Ablation.} Classification accuracy will decrease when $K$ is reduced and when $L_{2}$ or $L_{3}$ are omitted from the objective function.}\label{table:remove}
\end{subtable}
\newline
\vspace*{0.2 cm}
\newline
\begin{subtable}{\linewidth}\centering
{\begin{tabular}{c| c  c c |c c }
Fusion Method & TA\% $\uparrow$ & IT(s) $\uparrow$ & RS $\downarrow$  & LP & CT\\ \hline
GRU & $\bf{90.73}$ & $\bf{15.78}$ & $2.17$ & $0.94$ & $0.62$ \\
Average Pool & $86.83$ & $20.63$ & $\bf{2.02}$ & $0.88$ & $0.68$ \\
Max Pool & $83.90$& $17.01$ & $\bf{2.02}$ & $0.92$ & $0.66$ \\ 
\end{tabular}}
\caption{\textbf{Feature Fusion.} Classification accuracy will increase when GRU units are used to integrate multi-scale features.}\label{table:fusion}
\end{subtable}

\caption{Effect of individual model components on Total Accuracy (TA), Inference time (IT), Relative Size (RS), ratio of low resolution predictions (LP) and the calibrated confidence threshold (CT).}\label{table:ablation}
\end{table}

The aim of the SOS protocol is to achieve rapid WSI classification without: (a) substantially increasing model size; or (b) compromising on classification accuracy. Thus, we validate the effectiveness of our method by analysing quantitative metrics for classification accuracy, model size, and processing speed. The design specifications for the SOS protocol, such as the number of patches processed at high resolution ($K=10$), and the multi-image feature aggregation method (GRU), were determined by the outcomes of the ablation studies in Table \ref{table:ablation}.  As described in Section \ref{sect:related}, the existing approaches for WSI classification can be broadly grouped into: (a) Patch-Level; (b) Conventional Multi-Scale; and (c) Dynamic Multi-Scale methods. We assess the quantitative performance of our SOS protocol against each of these commonly used methods and additionally include image-level WSI classification performance for comparison (Tables \ref{table:final_results} and \ref{table:f1}). We also qualitatively analyze the outputs of our patch selection network to validate the effectiveness of our HRN (Figures \ref{fig:bad} and \ref{fig:patches}).

\subsection{Data Preprocessing}
For computational efficiency, all WSIs were preprocessed into $s$ and $\mathcal{S}$ prior to training and inference. To produce $s$ where $H=1000$, $W=1000$ and $C=1$; we downscale the full resolution WSIs by a factor of $40$. To produce $\mathcal{S}$, we segment full resolution WSIs into non-overlapping patches with the same dimensions as $s$; thus, the total number of patches per WSI is $1600$. Since ResNet18 \cite{resnet} was designed to be used on tricolor image inputs, we modified the first convolutional layer to have single channel inputs to be compatible with these monocolor images.

\begin{table}
\small
\begin{center}
\begin{tabular}{c| c  c  c  c|c}
Method & TA\% $\uparrow$ & RS $\downarrow$ & IT(s) $\downarrow$ & SB $\uparrow$ & LP\\ \hline
Image-Level & $81.95$ & $\bf{1.00}$ & $\bf{8.37}$ & $\bf{14.59}$&-\\
Patch-Level & $69.27$ &  $1.50$ &$94.10$ & $1.3$&-\\ 
Multi-Scale & $85.37$ &  $2.17$ &$122.10$ &  $1.00$&- \\ \hline
RDMS & $88.78$ &  $3.83$ &$57.30$ &  $2.13$ &0.55 \\
SOS (ours) & $\bf{90.73}$ &  $2.17$ & $15.78$ & $7.74$  &0.94\\
\end{tabular}
\end{center}
\caption{Comparison of Total Accuracy (TA), Relative Size (RS), Inference Time (IT) and Speed Boost (SB) metrics. The ratio of low resolution predictions (LP) is also provided for the dynamic multi-scale classification methods.}
\label{table:final_results}
\end{table}

\subsection{Method Comparison}
The details of each of the methods used for comparison are described below. All models were evaluated after training for $20$ epochs using a learning rate of $10^{-3}$. 

\textbf{Image-Level.} The Image-Level method performs classification on single-scale image-level features. Specifically, we train a ResNet18 model \cite{resnet} to directly compute the class label ($q$) from downscaled WSI ($s$) inputs. We optimize image-level classification by minimizing the cross entropy loss for ground truth and predicted class probabilities.

\textbf{Patch-Level.} The Patch-Level method performs classification on single-scale patch-level features. The network architecture is essentially the same as the proposed model; however, we remove the residual connection to image-level features. Thus, only high resolution features from the set of $K$ patches are used to classify the WSI. We optimize patch-level classification by minimizing the cross entropy loss for ground truth and predicted class probabilities.

\textbf{Conventional Multi-Scale.} The Conventional Multi-Scale method performs classification on multi-scale image-level and patch-level features. The design of the Multi-Scale method is similar to the Patch-Level network; however, we restore the residual connection to image-level features as shown in Figure \ref{fig:ionet}. Thus, features from both low resolution and high resolution image scales are used to classify each WSI. We optimize multi-scale classification by minimizing the cross entropy loss for ground truth and predicted class probabilities. In this paper, we refer to this method as Multi-Scale.

\textbf{Dynamic Multi-Scale.} The Dynamic Multi-Scale method is based on the Reinforced Auto-Zoom Net (RAZN) framework proposed by Dong \textit{et al}. \cite{razn}. Specifically, we remove the switch statement (Algorithm \ref{code:switch}) from the EPU and attach a policy network, based on the ResNet architecture, to the front end of our model to decide whether to use the LRN or the HRN for WSI classification. Since RAZN is designed for WSI segmentation tasks, we modify the reward function to be suitable for LKS classification as follows:

\begin{equation}
\label{eq:razn}
R(a) = a\frac{L_{ce_{2}} - L_{ce_{1}}}{L_{ce_{1}}},
\end{equation}

where $a \in \{0,1\}$, represents the policy action to use the LRN or HRN respectively. Thus, when $a = 1$ then the reward $R(a)$ is positive if $L_{ce_{2}} > L_{ce_{1}}$, and negative if $L_{ce_{2}} < L_{ce_{1}}$. We then optimize the network using the policy gradient method described in  \cite{razn}. We refer to this method as Reinforced Dynamic Multi-Scale (RDMS).

\subsection{Quantitative Evaluation}
\textbf{Processing Speed.} Processing speed is evaluated using Inference Time (IT) and Speed Boost (SB) metrics. The IT is a raw measurement of inference time (in seconds) on the test dataset. SB indicates the factor by which inference time is reduced relative to the baseline model. The model used as the baseline for processing speed is the conventional Multi-Scale model. The reason for selecting this model as the baseline is because it achieved the highest accuracy among the static classification models. Thus, it defines the baseline speed at which the highest accuracy can be achieved without adapting the resolution used to process each WSI.

\textbf{Model Size.} The model size is evaluated using the Relative Size (RS) metric. The RS metric indicates the relative increase in size compared to the simple image-level ResNet18 classifier used as the backbone in all our multi-scale methods e.g. an RS of $2$ indicates that the model has twice as many parameters as the simple ResNet18 classifier.

\textbf{Classification Accuracy.} The classification accuracy is evaluated using the Total Accuracy (TA) metric. However, since the majority of samples in the dataset are negative, the TA does not provide an adequate measure of classification performance on individual classes. To assess individual class performance, we provide the F1 score (F1), Precision (PR), Recall (RE), and Specificity (SP) for each of the four WSI classes in Table \ref{table:f1}.

\begin{table}
\small
\begin{subtable}{\linewidth}\centering
{\begin{tabular}{c| c  c  c  c}
Method & F1 $\uparrow$ & PR $\uparrow$ & RE $\uparrow$ & SP $\uparrow$ \\ \hline
Image-Level &$0.8800$&$0.8115$&$\bf{0.9612}$&$0.7745$ \\
Patch-Level&$0.7967$&$0.6853$&$0.9515$&$0.5588$ \\
Multi-Scale &$0.9083$&$0.8609$&$\bf{0.9612}$&$0.8431$ \\ \hline
RDMS &$0.9300$&$0.9588$&$0.9029$&$\bf{0.9608}$ \\
SOS (ours) & $\bf{0.9406}$ &$\bf{0.9596}$&$0.9223$&$\bf{0.9608}$ \\
\end{tabular}}
\caption{\textbf{Negative.} Evaluation of Negative classification performance.}\label{table:neg}
\end{subtable}

\begin{subtable}{\linewidth}\centering
{\begin{tabular}{c| c  c  c  c}
\\
Method & F1 $\uparrow$ & PR $\uparrow$ & RE $\uparrow$ & SP $\uparrow$ \\ \hline
Image-Level &$0.8989$&$0.9090$&$0.8889$&$\bf{0.9750} $\\
Patch-Level &$0.8471$&$0.9000$&$0.8000$&$\bf{0.9750}$ \\
Multi-Scale &$0.8706$&$\bf{0.9250}$&$0.8222$&$0.9813$ \\ \hline
RDMS &$0.9149$&$0.8776$&$\bf{0.9556}$&$0.9625$ \\
SOS (ours) &$\bf{0.9348}$&$0.9149$&$\bf{0.9556}$&$\bf{0.9750}$ \\
\end{tabular}}
\caption{\textbf{AMA.} Evaluation of AMA classification performance.}\label{table:ama}
\end{subtable}

\begin{subtable}{\linewidth}\centering
{\begin{tabular}{c| c  c  c  c}
\\
Method & F1 $\uparrow$ & PR $\uparrow$ & RE $\uparrow$ & SP $\uparrow$ \\ \hline
Image-Level &$0.6667$&$0.7368$&$0.6087$&$0.9371$ \\
Patch-Level &$0.2353$&$0.3636$&$0.1739$&$0.912$ \\
Multi-Scale &$0.7778$&$0.7955$&$0.7609$&$\bf{0.9434} $\\ \hline
RDMS &$0.8367$&$0.7885$&$\bf{0.8913}$&$0.9308$ \\
SOS (ours) &$\bf{0.8542}$&$\bf{0.8200}$&$\bf{0.8913}$&$\bf{0.9434}$ \\
\end{tabular}}
\caption{\textbf{SMA-V.} Evaluation of SMA-V classification performance.}\label{table:smav}
\end{subtable}

\begin{subtable}{\linewidth}\centering
{\begin{tabular}{c| c  c  c  c}
\\
Method & F1 $\uparrow$ & PR $\uparrow$ & RE $\uparrow$ & SP $\uparrow$ \\ \hline
Image-Level &$0.1667$&$\bf{1.000}$&$0.0909$&$\bf{1.000} $\\
Patch-Level &$0.0000$&$0.0000$&$0.000$&$\bf{1.000}$ \\
Multi-Scale &$0.4706$&$0.6667$&$0.3636$&$0.9897$ \\ \hline
RDMS &$0.5556$&$0.7143$&$0.4545$&$0.9897$ \\
SOS (ours) &$\bf{0.7000}$&$0.7778$&$\bf{0.6364}$&$0.9897$ \\
\end{tabular}}
\caption{\textbf{SMA-T.} Evaluation of SMA-T classification performance.}\label{table:smat}
\end{subtable}
\caption{Evaluation of F1 scores, Precision (PR), Recall (RE) and Specificity (SP) for each of the four WSI classes.}
\label{table:f1}
\end{table}

\section{Discussion}

\begin{figure*}[t]
\begin{center}
   \includegraphics[width=\linewidth]{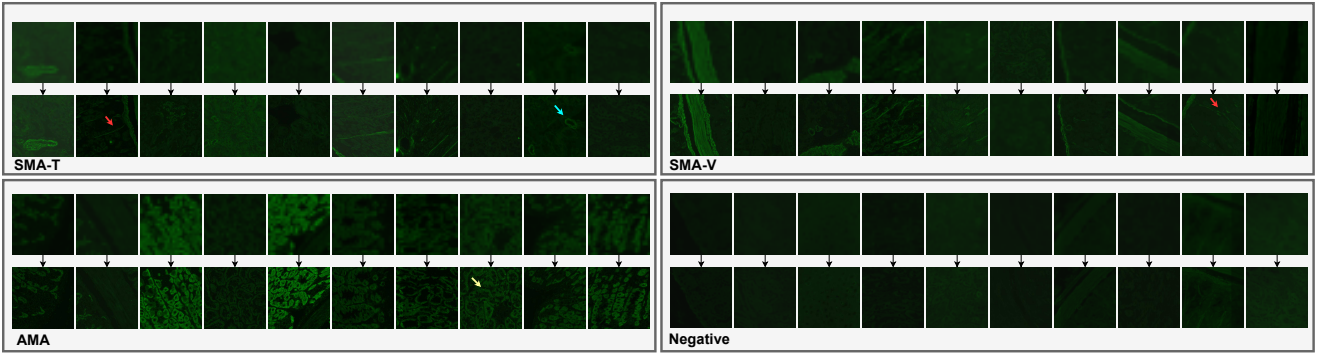}
\end{center}
   \caption{Examples of WSI patch regions sampled by the HRN for each of the WSI classes. The sampled patch regions are provided in low (top) and high (bottom) resolution to compare the different levels of detail at each scale. The colored arrows indicate examples where diagnostically significant features could not be resolved at the lower resolution. At the LRN input resolution, it is difficult to observe the fine granular staining of mitochondria (yellow arrow) and virtually impossible to resolve the staining of smooth muscle actin fibres in the stomach (red arrow) and kidney (blue arrow). However, in the corresponding high resolution patches sampled by the HRN, these distinguishing features are clearly visible.}
\label{fig:patches}
\end{figure*}

The results in Table \ref{table:final_results} indicate that our proposed method was able to reduce inference time by a factor of $7.74$ while improving classification accuracy compared to the conventional multi-scale approach. The improved classification accuracy was unexpected because the multi-scale network processes low resolution contextual features and high resolution local features for every WSI while our network only uses single-scale low resolution features for the majority of test samples. We suspect that training our model to use LRN features for patch selection and classification may explain the improved accuracy as it has been shown that jointly learning the tasks of detection and classification have a beneficial effect on model performance \cite{joint_task}. 

The RDMS method was also shown to improve accuracy and reduce inference time. However, as shown in Table \ref{table:final_results}, adding the additional policy network has introduced significantly more parameters than our proposed method. The limitations of using RDMS for classification tasks are demonstrated clearly by the LP of 0.55; meaning only $55\%$ of samples in the test dataset were classified at low resolution compared to our $94\%$. The fact that our model still achieved a higher accuracy when processing fewer samples at high resolution indicates that RDMS is using the HRN excessively. This behavior was expected because the policy reward in RDMS (Equation \ref{eq:razn}) is always zero unless the ``zoom" (HRN processing) action is sampled. This means there is an incentive to utilize HRN processing whenever the incurred loss is lower than the LRN prediction i.e. $L_{ce_{2}} < L_{ce_{1}}$. However, in classification tasks, a lower cross entropy loss is not a perfect estimate of classification accuracy because the LRN may still reliably predict the correct class label. In our method, we encourage the network to maximize the use of LRN for classification in these cases by minimizing $L_{he}$ (Equation \ref{eq:hesitation}). Thus, we could classify WSIs significantly faster than RDMS without compromising on classification accuracy.

Figure \ref{fig:patches} provides examples of sampled patches for each of the WSI classes. At low resolution, it is virtually impossible to resolve the fine smooth muscle actin fibres in SMA-T and SMA-V patches. The staining of actin fibres is an essential diagnostic feature of smooth muscle antibodies \cite{ald}. The inability to resolve these fibres at low resolution likely explains why the F1 scores for these classes are substantially lower with the Image-Level classifier than with the Multi-Scale, RDMS and SOS methods; which all incorporate high resolution WSI features (Table \ref{table:f1}). The Patch-Level method also has access to the high resolution WSI features; however, without any spatial context features from the low resolution WSI, it obtained the lowest classification accuracy of all tested methods (Table \ref{table:final_results}). 


While the HRN clearly improves the accuracy of our model, the hard patch attention mechanism often selects patches containing no tissue at all (Figure \ref{fig:bad}). The selection of undesirable patches likely explains why average and max pooling methods do not perform as well as the gated recurrent unit for aggregating multi-image features (Table \ref{table:ablation}) --- as recurrent neural networks are known to be more robust to noisy WSI patches \cite{patch_lstm}. 

From the ablation studies in Table \ref{table:ablation}, it is clear that reducing the number of patches processed by the HRN will have an adverse effect on classification accuracy. Omitting the paradoxical loss ($L_{2}$) during training also results in a significant drop in classification accuracy. The paradoxical loss is used to prevent overconfident estimates by the LRN; without it, the model becomes biased towards classifying examples at low resolution (as indicated by the LP of $0.97$). A smaller reduction in classification accuracy is also observed when the executive loss term ($L_{3}$) was omitted during training. The executive loss term helps regulate the network to regress class probability estimates on the appropriate side of the decision boundary. Without the executive loss, class probabilities can be estimated without feedback on how they affected EPU decisions; which can result in the suboptimal image scale being selected to classify a WSI.  

\section{Conclusion}

In this paper, we show that it is possible to reduce inference time in WSI classification tasks, without compromising on accuracy, by restricting high resolution patch-based processing to cases that cannot be classified confidently at low resolution. The effectiveness of our proposed SOS protocol is demonstrated on the LKS dataset; which presents the challenging task of finding the optimal trade-off between low resolution WSI processing and patch-based processing. To evaluate the generalizability of the proposed framework, in future works, we will perform experiments on a variety of IIF WSI datasets that are being assembled in collaboration with Sullivan Nicolaides Pathology.

\section{Acknowledgements}
This project has been funded by Sullivan Nicolaides Pathology and the Australian Research Council (ARC) Linkage Project [Grant number LP160101797].


\end{document}